\documentclass[pmlr]{jmlr}% new name PMLR (Proceedings of Machine Learning)

\RequirePackage{graphicx}
 \usepackage{booktabs}
\usepackage{longtable}% for long tables
\makeatletter
\def\set@curr@file#1{\def\@curr@file{#1}} %temp workaround for 2019 latex release
\makeatother
\usepackage[load-configurations=version-1]{siunitx} % newer version

\theorembodyfont{\upshape}
\theoremheaderfont{\scshape}
\theorempostheader{:}
\theoremsep{\newline}

\jmlrvolume{298}
\jmlryear{2025}
\jmlrworkshop{Machine Learning for Healthcare}

\title[Learning to Call]{Learning to Call: A Field Trial of a Collaborative Bandit Algorithm for Optimizing Call Timing in Mobile Maternal Health}

\author{\Name{Arpan Dasgupta*}
\Email{arpandg@google.com}\\
\addr Google Deepmind\\ 
\Name{Mizhaan Maniyar*}
\Email{mizhaan@google.com}\\
\addr Google Deepmind\\
\Name{Awadhesh Srivastava}
\Email{awadhesh@armman.org}\\
\addr ARMMAN\\ 
\Name{Sanat Kumar}
\Email{sanat@armman.org}\\
\addr ARMMAN\\
\Name{Amrita Mahale}
\Email{amrita@armman.org}\\
\addr ARMMAN\\
\Name{Aparna Hegde}
\Email{aparnahegde@armman.org}\\
\addr ARMMAN\\
\Name{Arun Suggula}
\Email{arunss@google.com}\\
\addr Google Deepmind\\
\Name{Karthikeyan Shanmugam}
\Email{karthikeyanvs@google.com}\\
\addr Google Deepmind\\ 
\Name{Milind Tambe}
\Email{milindtambe@google.com}\\
\addr Google Deepmind\\ 
\Name{Aparna Taneja}
\Email{aparnataneja@google.com}\\
\addr Google Deepmind\\
}

\begin{document}

\maketitle

%%%%%%%%%%%%%%%%%%%%%%%%%%%%%%%%%%%%%%%%%%%%%%%%%%%%%%%%%%%%%%%%%%%%%%%%

\begin{abstract}
    Mobile health (mHealth) programs  utilize automated voice messages to deliver health information, particularly targeting underserved communities, demonstrating the effectiveness of using mobile technology to disseminate crucial health information to these populations, improving health outcomes through increased awareness and behavioral change. India's Kilkari program delivers vital maternal health information via weekly voice calls to millions of mothers. However, the current random call scheduling often results in missed calls and reduced message delivery. This study presents a field trial of a collaborative bandit algorithm designed to optimize call timing by learning individual mothers' preferred call times. We deployed the algorithm with around $6500$ Kilkari participants as a pilot study, comparing its performance to the baseline random calling approach. Our results demonstrate a statistically significant improvement in call pick-up rates with the bandit algorithm, indicating its potential to enhance message delivery and impact millions of mothers across India. This research highlights the efficacy of personalized scheduling in mobile health interventions and underscores the potential of machine learning to improve maternal health outreach at scale.
\end{abstract}

\section{Introduction}

Maternal health remains a critical public health concern in India and various other developing countries globally (\cite{momconnect}, \cite{nshimirimana2012designing}, \cite{hategeka2019effect}, \cite{Ward2020-tq}, \cite{mHealthTanzania2013}), with millions of women having limited access to timely and accurate information during pregnancy and postpartum. Mobile health (mHealth) programs (\cite{Murthy2020}, \cite{Chowdhury2019}, \cite{Kabukye2021-yy}) that use automated voice messages, have the ability to deliver such critical maternal and child health information. Recognizing the need for information and the promise of mHealth programs, the Government of India launched the Kilkari program, a nationwide mobile health initiative that delivers weekly voice messages that contain essential maternal health information to more than 10 million registered mothers~\footnote{\url{https://rchrpt.mohfw.gov.in/RCHRPT/Kilkari/Kilkari_Message.aspx}}. mHealth programs such as these play a vital role in reducing maternal mortality rates - a key target within the WHO's Sustainable Development Goals \cite{WHOTargetMaternal}. These messages cover vital topics such as iron and calcium supplementation, antenatal care, and postnatal practices, aiming to improve maternal health outcomes throughout the country.

However, the effectiveness of this large-scale program is contingent upon successful message delivery. Currently, Kilkari employs a random call scheduling strategy, attempting to reach mothers, with up to nine re-attempts (until the call is picked up), but without considering individual preferences for call timing. This approach often results in missed calls, crucial bandwidth spent on re-attempts and most importantly limiting the reach and impact of crucial health information \cite{JJH, mohan2021can, lalan2023analyzing}. To address this challenge, recently, a stochastic bandit approach was proposed to learn appropriate timing for calls to mothers. Whereas learning individually the appropriate timing to call each mother is expensive, a collaborative bandit approach attempts to harness similarity among the mothers to jointly learn their preferences for call timings \cite{pal2024improving}. Whereas this approach has shown promise in simulations, its performance in real-world field trials remains unknown. To address this limitation, this paper presents a field trial of the collaborative bandit algorithm \cite{pal2024improving} designed to optimize call scheduling by learning mothers' preferred call times.

\subsection*{Generalizable insights}
Collaborative bandit algorithms offer a promising approach for personalized intervention delivery in mobile health. By iteratively learning from user responses and interactions, these algorithms can adapt to individual preferences and maximize engagement. In this study, we implemented a collaborative bandit algorithm within the Kilkari platform and conducted a field trial involving approximately $6500$ beneficiaries. Our goal was to evaluate the algorithm's ability to improve call pick-up rates compared to the baseline random calling strategy.

This research contributes to the growing body of literature in the application of machine learning in mobile health interventions \cite{verma2023deployed, mate2022field, nair2022adviser}. By demonstrating the effectiveness of a collaborative bandit algorithm in a real-world setting, we highlight the potential for personalized call scheduling to enhance the reach and impact of maternal health programs at scale. Given the national scope of Kilkari and the potential for improved message delivery to millions of beneficiaries, our findings have significant implications for public health policy and practice in India and beyond.

\subsection*{Key note on the experiments reported in this paper} This work was conducted as a joint effort between a research team from a non-profit in India called ARMMAN (\cite{armmanArmmanHome}) and Google Deepmind India a non-profit organization in India as reflected in the co-authorship of this paper.
It is crucial to highlight that the beneficiary data utilized in this research is fully anonymized, and no socio-demographic features were available to the research team. To ensure data privacy and security, the experimental infrastructure was managed exclusively by the ARMMAN team, who were the only individuals with access to the raw beneficiary data.
The Google Deepmind researchers contributed by advising the ARMMAN team on the collaborative bandit algorithm, specifically the algorithm’s implementation and subsequently collaborating on the analysis of the resulting study. The ARMMAN team has followed general guidelines related to ethics approvals laid down by Indian Council for Medical Research (ICMR).

%%%%%%%%%%%%%%%%%%%%%%%%%%%%%%%%%%%%%%%%%%%%%%%%%%%%%%%%%%%%%%%%%%%%%%%%

\section{Related Work}

% \paragraph{AI in mHealth Programs}
mHealth programs provide essential health information though automated voice messages to a large number of beneficiaries \cite{hegde2016assessing, murthy2020effects}, which implies any improvement to the program positively affects a lot of beneficiaries. Previously, AI has been applied to schedule interventions \cite{mate2022field, verma2023deployed} and showed positive behavioral outcomes \cite{dasgupta2024preliminary}. 
\citet{lefevre2019stage} talk about the protocol for an individually controlled randomized control trial in an attempt to show the effectiveness of Kilkari.

Scheduling the time of the day the beneficiaries are called using collaborative bandits \cite{pal2024improving} showed promise in simulation and this work aims to test it out in a pilot study. The analysis in \cite{bashingwa2021assessing} shows that there are preferred slots for calling beneficiaries. They further show that most calls which are picked are done so by the third attempt. These factors point towards the advantages of scheduling these calls in a non-random manner.

% \paragraph{Collaborative Bandits}
Multi-armed bandits represent a well-researched and potent approach for tackling diverse resource allocation challenges. Numerous methodologies, including phased elimination \cite{lattimore2020bandit, slivkins2019introduction}, Upper Confidence Bound (UCB) \cite{auer2002finite}, Thompson Sampling \cite{thompson1933likelihood, agrawal2012analysis}, and Best-arm Identification \cite{agrawal2020optimal, garivier2016optimal}, have undergone thorough investigation. The collaborative bandit problem has witnessed a surge in interest recently, driven by the widespread adoption of recommender systems \cite{bresler2016collaborative, dadkhahi2018alternating}. Under specific conditions, several algorithms with robust theoretical guarantees have been developed \cite{pal2023optimal, jain2022online}. An algorithm suited for scenarios with approximate low-rank structure, was introduced by \cite{pal2024improving} and is evaluated in this field study.

%%%%%%%%%%%%%%%%%%%%%%%%%%%%%%%%%%%%%%%%%%%%%%%%%%%%%%%%%%%%%%%%%%%%%%%%

\section{Background}

\subsection{Kilkari}

\begin{figure}
    \centering
    \includegraphics[width=0.4\textwidth]{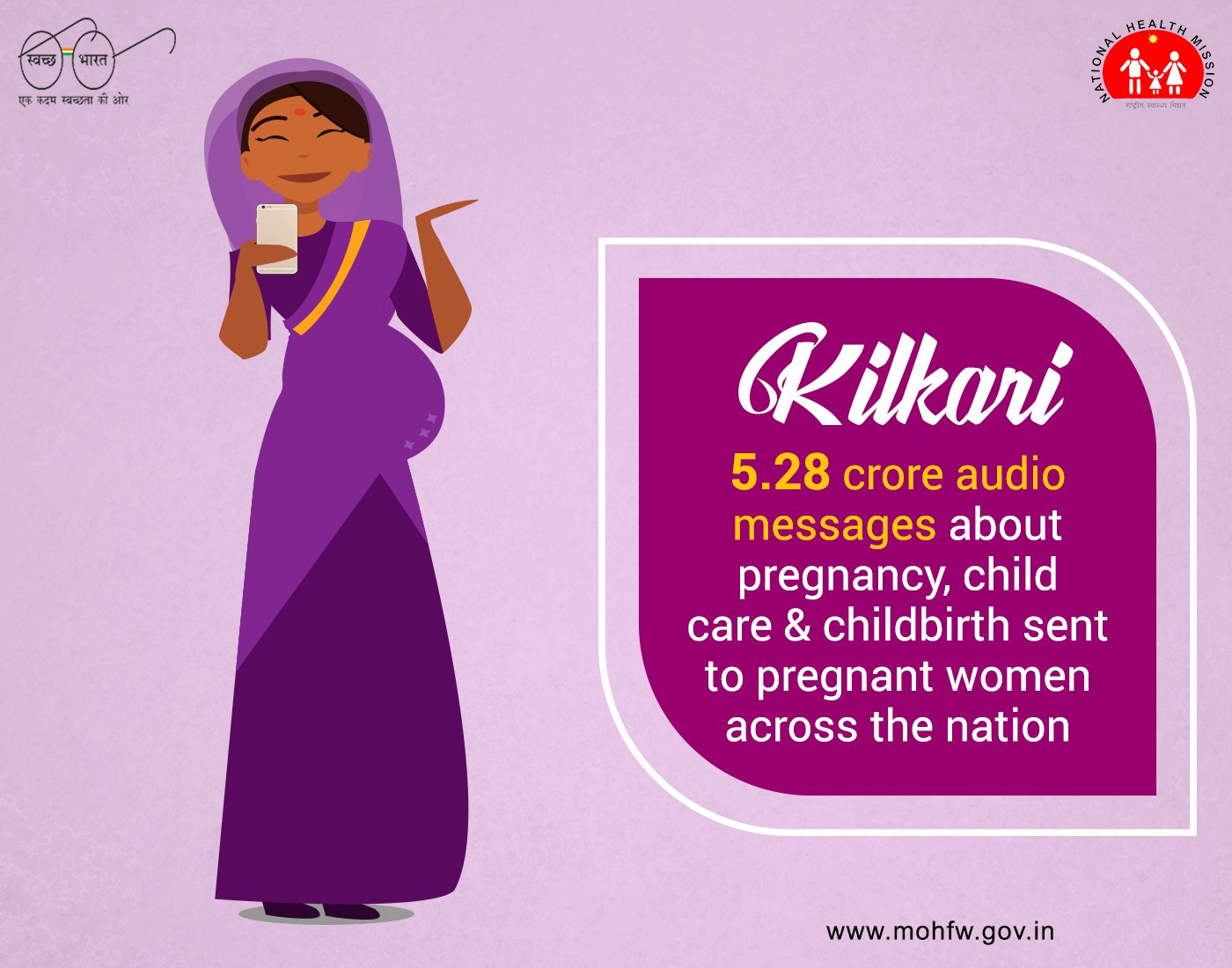}
    \caption{An example poster image from the official X account of the MOFHW. Link \href{https://x.com/MoHFW_INDIA/status/979600610952654848/photo/1}{here}.}
    \label{fig:kilkari_poster}
\end{figure}

Kilkari %\cite{ARMMAN_kilkari} 
is the world’s largest mobile health program focused on maternal and child health  (Figure \ref{fig:kilkari_poster}).  It is conducted by India's Ministry of Health and Family Welfare in partnership with an NGO in India (name of NGO withheld for the sake of anonymity).   Kilkari uses pre-recorded voice calls to deliver vital preventive care information on maternal and infant health to pregnant women and new mothers. The program aims to improve access to healthcare information for pregnant women, mothers of infants, and their families, particularly in underserved communities. 

However, these programs face challenges, including limited beneficiary phone access and unknown time preferences, which hinder timely outreach and lead to poor engagement.  Specifically, low listenership of the automated voice messages is a major challenge.  Even with multiple call attempts, approximately $23\%$ of beneficiaries are not reached.  Consistent low listenership can even lead to beneficiaries being dropped from the program, which can occur if beneficiaries listen to less than $25\%$ of the messages for six weeks in a row.

To address the challenge of low listenership, the use of a stochastic bandit approach could be very useful to learn the favored time slot of individual mothers/beneficiaries.  This is important because factors such as limited phone access, working hours, and household responsibilities significantly affect the likelihood of answering a call at a given time slot.  By quickly identifying good time slots for each beneficiary, engagement with the calls can be improved, and beneficiaries can be retained in the program.  Furthermore, optimizing the time slot to send automated voice messages can help to reduce automatic dropouts and save bandwidth.
\subsection{Collaborative Bandits}
\citet{pal2024improving} address the challenge of optimizing time slot selection in mobile health programs like Kilkari, where the goal is to deliver automated voice messages to beneficiaries at times they are most likely to engage. To tackle this, \cite{pal2024improving} formulate the problem as a multi-agent multi-armed bandit problem \cite{slivkins2019introduction}. Here, each beneficiary is modeled as an agent, and each possible time slot for delivering the message is considered an arm. The key idea is to learn the preferences of these agents (beneficiaries) for different arms (time slots) through repeated interactions (i.e., attempting to deliver messages). 

To efficiently solve this multi-agent bandit problem, the underlying algorithm in \cite{pal2023optimalalgorithmslatentbandits, pal2023optimal} is applied to the current problem. This framework leverages the assumption that the preferences of beneficiaries are not entirely independent but rather share some underlying structure. Specifically, it assumes that the matrix representing beneficiaries' preferences (e.g., the probability of a beneficiary picking up a call at a given time slot) is approximately low-rank. This low-rank assumption implies that there are a few latent factors that explain a significant portion of the variability in beneficiaries' time slot preferences. The collaborative bandit algorithm exploits this low-rank structure to learn more efficiently by sharing information across beneficiaries, rather than learning each beneficiary's preferences in isolation.

\citet{pal2024improving} introduces two novel algorithms: Greedy Matrix Completion (MC) and Phased MC. Phased MC is the key algorithm we use in this work. Greedy MC first has a long random phase where arms are picked randomly, followed by prediction which is followed thereafter. Phased MC operates by updating the estimates in "phases" which implies that no lengthy exploration phase is required to obtain an initial estimate. To allow for exploration during prediction, a Boltzmann noise \cite{cesa2017boltzmann} is added. It also uses variance reduction techniques to improve robustness to noise. Since the exploration phase required in this algorithm is not as large, it hence prevents chances of dropout early in the trial due to random calls.

%%%%%%%%%%%%%%%%%%%%%%%%%%%%%%%%%%%%%%%%%%%%%%%%%%%%%%%%%%%%%%%%%%%%%%%%

\section{Experimental Design}

This field trial was conducted in the Kalahandi and Puri districts of Odisha, India upon the guidance of the ARMMAN team, to evaluate the effectiveness of the collaborative bandit algorithm in improving call pick-up rates within the Kilkari maternal health program.
\newline \subsection{Randomization}
Beneficiaries were randomly assigned to either the Random or Treatment group to minimize bias and ensure comparability between the groups. We consider only those users that didn't drop out \textit{after} the baseline phase as mentioned in \ref{subsec:trial_phases} to ensure a fair analysis.
%Beneficiaries (mothers) were randomly assigned to one of two groups, \\
\newline\textbf{Random Group (Control)}: This group received calls using the current Kilkari \textit{call retry algorithm} \cite{Bashingwa2021}. This group considered $6416$ beneficiaries for the analysis. 
\newline
\textbf{Treatment Group (Collaborative Bandit)}: This group received calls scheduled using a collaborative bandit algorithm designed to learn and adapt to individual beneficiaries' preferred call times. This group considered $6490$ beneficiaries for the analysis.
The algorithm from \cite{pal2024improving} relies on an offline low rank
matrix completion oracle $\mathcal{O}$. For an unknown matrix $Z \in \mathbb{R}^{m \times n}$, $\mathcal{O}$ takes a subset of noisy observations of a matrix as position $\omega$ ($\{M_{ij}\}_{(i, j) \in \omega}$) as input, and returns an estimate $ \hat{Z}$ of $Z$. To implement this oracle, we minimize the following nuclear norm regularized objective:
\begin{align}
    \textrm{minimize}_{\hat{Z}} \sum_{(i, j) \in \omega} (M_{ij} - \hat{Z}_{ij})^2 + \lambda || \hat{Z} ||_{\star},
\end{align}
where $\lambda > 0$ is the regularization parameter and $|| \cdot ||_{\star}$ denotes the nuclear norm for a matrix as shown in \cite{pal2024improving}. The hyper-parameters of the model were tuned by holding out 20\% of the most recent call data as validation, and then choosing the best hyper-parameters by calculating the score on this data based on predictions via grid search. 
% Using this method, we get a value $\lambda = 0.9$.

\subsection{Trial Phases}
\label{subsec:trial_phases}
The trial consisted of two distinct phases: \newline (i) \textbf{Baseline Phase (Weeks 1-3) [7th January - 26th January, 2025]}:
Both the Random and Treatment groups received calls using the standard Kilkari random calling strategy.
This phase served to establish a baseline for call pick-up rates and to collect data for the collaborative bandit algorithm in the Treatment group to initiate learning beneficiaries’ preferences.
The exact number of calls attempted for each beneficiary during this phase will be detailed in Section \ref{sec:results}.
\newline(ii) \textbf{Intervention Phase (Weeks 4-5) [27th January - 9th February 2025]}:
The Random group continued to receive calls using the random calling strategy.
The Treatment group, however, received calls scheduled based on the preferences learned by the collaborative bandit algorithm during the Baseline Phase in an iterative manner.
The exact number of calls attempted for each beneficiary during this phase will be detailed in Section \ref{sec:results}.\\
\subsection{Data Collection}
Call logs were collected for all beneficiaries in both groups, recording the date, time, and outcome (answered/missed) of each call attempt.
The total number of beneficiaries in each group will be noted in the Section \ref{sec:results}.\\
\subsection{Outcome Measure}
The primary outcome measure was the call pick-up rate, defined as the proportion of successful call pick-ups out of the total number of call attempts, for each group during the Intervention Phase.
\newline
\textbf{Ethical Considerations:}
% todo
No ethical approvals were required for this study as it was deployed on an existing program and counts as a program improvement.\\
\textbf{Statistical Analysis:}
Statistical analysis was conducted to compare the call pick-up rates between the Random and Treatment groups during the Intervention Phase. We used a simple two-sample t-statistic to verify the statistical significance between the two groups, across the baseline and intervention phase. More information can be found in the next section.
%%%%%%%%%%%%%%%%%%%%%%%%%%%%%%%%%%%%%%%%%%%%%%%%%%%%%%%%%%%%%%%%%%%%%%%%
\section{Preliminaries}\label{sec:preliminaries}

In this section we mathematically define the call pick-up rates and its variants that are used for the analysis in the later section. The index $i$ represents a beneficiary (or user), $j \in [1, 7]$ being one of the seven time slot IDs chosen, $t$ being the day, and $r \in [0, 2]$ being the re-attempt number for that slot, i.e. $r=0$ being the first call, and $r=1$ being the second call made, if the first call wasn't picked up. The time windows for the 7 time slot IDs are detailed in the Appendix in Table \ref{tab:timeslot_ids_details}. Let $call$ be mapped uniquely to the tuple $(i, j, t, r)$. Let $A_{call} \equiv A_{i,j,t,r} \in \{0, 1\}$ denote whether a call attempt was made for user $i$ during time-slot $j$ on day $t$ and whether it was the $r$-th re-attempt. Similarly, let $p_{call} \equiv p_{i, j, t, r} \in \{ 0, 1 \}$ denote whether an attempted call was picked or not. We assume that the set of calls $\{ call | p_{call} = 1 \} \subseteq \{ call | A_{call} = 1\} $. We now define the pooled pick-up rate as,
\begin{align}
\label{eq:pooled_pr}
    PR_{pooled} = \frac{\sum_{\forall i,j,t,r} p_{i, j, t, r}}{\sum_{\forall i, j, t, r} A_{i, j, t, r}} \equiv \frac{\sum_{\forall call} p_{call}}{\sum_{\forall call} A_{call}}.
\end{align}
Alternatively, we can define a user-specific pick-up rate (PR) and its average as such
\begin{align}
\label{eq:user_pr}
    PR_{i} = \frac{\sum_{\forall j, t, r} p_{i, j, t, r}}{\sum_{\forall j, t, r} A_{i, j, t, r}}, & \qquad
    PR_{user} = \frac{\sum_{\forall i} PR_{i}}{\sum_{\forall i} 1}.
\end{align}
The metric $PR_i$ can be seen as an estimate of the probability of user $i$ picking up a call.

\section{Results} 
\label{sec:results}

% \textit{Using data from 27th January onwards for intervention}
In this section we analyse the pick-up rates obtained for the two groups, i.e. treatment and control during the two phases. We present the pooled pick-up rate values obtain from \ref{eq:pooled_pr} in Table \ref{tab:overall_pickup_rates_active}. We notice that during the baseline phase the performance was similar across the two groups using a t-test.  Furthermore the difference between the two groups during the intervention phase was statistically significant.

%Furthermore, the reduction in performance between the two phases for the treatment group is of lesser significance (p-value $> 0.05$) as compared to the control group (p-value $<< 0.05$). Furthermore, the performance difference in the treatment group is much better and significant as compared to the control group in the intervention phase.
% \begin{table}[]
%     \centering
%     \caption{Pooled call pickup rates across all calls made in the respective phases.}
%     \label{tab:overall_pickup_rates}
% \begin{tabular}{p{1.5cm}|p{1.5cm}|p{1.5cm}|p{1.5cm}}
% 	\hline
% 	Group & $PR_{pooled}$ (baseline) & $PR_{pooled}$ (intervention) &  p-value \\ \hline
%   	Treatment  & 0.443 & \textbf{0.463} &\textbf{4.73e-07} \\
%     Control & 0.440 & 0.448 & 0.0428 \\ \hline
%   	p-value &  0.3369 &\textbf{0.0006} &  - \\ \hline
% \end{tabular}
% \end{table}

\begin{table}[]
    \centering
    \caption{Pooled call pick-up rates across calls made to only those users who didn't drop out in the intervention phase, i.e. active users.}
    \label{tab:overall_pickup_rates_active}
\begin{tabular}{l|r|r}
	\hline
	Group & $PR^{active}_{pooled}$ (baseline) & $PR^{active}_{pooled}$ (intervention) \\ \hline
   	Treatment  & 0.470 & 0.463 \\
    Control & 0.465 & 0.448 \\ \hline
   	p-value &  0.1345 & 0.0006 \\ \hline
\end{tabular}
\end{table}

% \begin{table}[]
%     \centering
%     \caption{Pooled call pickup rates across calls made to only those users who didn't drop out in the intervention phase, i.e. active users.}
%     \label{tab:overall_pickup_rates_active}
% \begin{tabular}{l|r|r|r|r}
% 	\hline
% 	Group & $PR^{active}_{pooled}$ (baseline) & $PR^{active}_{pooled}$ (intervention) & \% reduction & p-value \\ \hline
%   	Treatment  & 0.470 & 0.463 &  \textbf{-1.52}\% & 0.0849 \\
%     Control & 0.465 & 0.448 & -3.62 \% & 4.65e-05 \\ \hline
%   	p-value &  0.1345 & 0.0006 & - &  - \\ \hline
% \end{tabular}
% \end{table}

We now dive deeper in analyzing the difference that arose in the intervention phase, i.e. from the 27th of January to 9th February, comparing the pick-up rates in the treatment group with the control group. In order to remove outliers, we segregate beneficiaries with very high $PR_{i}$, i.e. those who always pick-up their calls and very low $PR_{i}$, i.e. those who never pick-up their calls. For the treatment group, we have 40.59\% users with a $PR_{i} = 1$ and 6.56\% users with $PR_{i} = 0$. While the control group has values of 38.46\% and 6.99\% respectively. In order to maintain a fair comparison, we remove the same fraction of users from both these groups, i.e. removing the top 40.59\% ($\max\{40.59\%, 38.46\%\}$) and bottom 6.99\% ($\max\{6.56\%, 6.99\%\}$) from both the groups according to their pick-up rate probability, i.e. $PR_{i}$ obtained via \ref{eq:user_pr}. We call these tiers, High Tier, Mid Tier and Low Tier, respectively, emphasising on the Mid Tier for most of the analysis results.

\subsection{Call Volumes}

Table \ref{tab:call_volumes} summarizes the number of calls made to each arm within each tier for the intervention phase only.

\begin{table}[h]
    \centering
    \caption{Call Volumes by Tier and Arm}
    \label{tab:call_volumes}
    \begin{tabular}{l|rr} % Add vertical lines
        \hline % Add horizontal lines
        Tier & Treatment & Control \\
        \hline % Add horizontal lines
        High & 5077 & 5222 \\
        Mid & 16775 & 17345 \\
        Low & 2789 & 2542 \\
        \hline % Add horizontal lines
    \end{tabular}
\end{table}

\subsection{Call Pick-up Rates}

Table \ref{tab:pickup_rates} presents the call pick-up rates for each arm within each tier, along with the corresponding p-values for statistical significance. We use a 2-sample t-test for the two arms in each of the 3 tiers and obtain the p-value according to the methodology mentioned in Section \ref{sec:preliminaries}.

\begin{table}[h]
    \centering
    \caption{Call Pick-up Rates by Tier and Arm}
    \label{tab:pickup_rates}
    \begin{tabular}{l|r|r|r|r} % Add vertical lines, use 'l' for text p-value
        \hline % Add horizontal lines
        Tier & Treatment & Control & \% improvement & p-value \\
        \hline % Add horizontal lines
        High & 1.0000 & 0.9732 & 2.75\% & 4.66e-32 \\
        Mid & 0.3763 & 0.3555 & 5.83\% & 7.07e-05 \\
        Low & 0.0100 & 0.0000 & NaN & 3.98e-07 \\
        \hline % Add horizontal lines
    \end{tabular}
\end{table}

\subsubsection*{High Tier}

By construction, the treatment group will have all its users with a $PR_{i} = 1, \forall i$, while the control group having an average, i.e ${PR_{user}} < 1$ representing the mean for this tier only.
% Both Arm1 and Arm3 exhibited a call pickup rate of 1.0000 (100\%) in Tier 1, indicating that beneficiaries in this tier consistently answered calls. As expected, there was no statistically significant difference between the arms in this tier (p = NS).

\subsubsection*{Mid Tier}

The treatment group achieved a call pick-up rate of 0.3763 (37.63\%), while the control group achieved a rate of 0.3555 (35.55\%). This difference amounts to a 5.83\% improvement, and it was statistically significant (p = 7.07e-05), demonstrating that the collaborative bandit algorithm significantly improved call pick-up rates for beneficiaries in the middle tier.

\subsubsection*{Low Tier}

For the Low Tier, the treatment group had a call pick-up rate of 0.01 (1.00\%), and the control group had a rate of 0.0000, as expected by construction, similar with the High Tier.

\subsection{Time slot wise analysis}

In order to see which time slots saw the most improvement, we analyse the calls made for the mid tier group (Tier 2) in Table \ref{tab:timeslot_pickup_rates}.
% and \ref{tab:timeslot_listenership}. 
% In Table \ref{tab:timeslot_listenership} we present the results for listenership, i.e. the fraction of the total call time the user was on call. For instance, if a user cuts a call that was supposed to last for 5 mins after 2 mins, then the listenership for that call would be 0.4. This metric gives us a more continuous insight along with the pickup rate as discussed. More formally, if $l_{call}$ is the listenership for a given call, $t^{on}_{call}$ is the time for which the user stayed on call, and $T_{call}$ is the total time duration of the call such that $0 \le t^{on}_{call} \le T_{call}$, then
% \begin{align}
% \label{eq:listenership}
%     l_{call} = \frac{t^{on}_{call}}{T_{call}} &, \qquad
%     L_{pooled} = \frac{\sum_{\forall call} l_{call}}{\sum_{\forall call} A_{call}}.
% \end{align}
However for each time slot, we use the following formulae for Tables \ref{tab:timeslot_pickup_rates}, 
% and \ref{tab:timeslot_listenership},
\begin{align}
\label{eq:timeslot_pooled}
    PR^{j}_{pooled} = \frac{\sum_{\forall i, t, r} p_{i, j, t, r}}{\sum_{\forall i, t, r} A_{i, j, t, r}}. 
    % & \qquad 
    % L^{j}_{pooled} = \frac{\sum_{\forall i, t, r} l_{i, j, t, r}}{\sum_{\forall i, t, r} A_{i, j, t, r}},
\end{align}
\begin{table}[h]
    \centering
    \caption{Pooled call pick-up rates $PR^{j}_{pooled}$ across all calls made in the respective time slot $j$ given by \ref{eq:timeslot_pooled}.}
    \label{tab:timeslot_pickup_rates}
    \begin{tabular}{c|r|r|r|r}
    	\hline
    	Time Slot ID & Treatment & Control & \% pick-up rate & p-value \\ \hline
       	1  & 0.3584 & 0.3337 & 7.4104 & \textbf{0.0563} \\
        2  & 0.3510 & 0.3365 & 4.2896 & 0.2695 \\ 
        3  & 0.3908 & 0.3625 & 7.8111 & \textbf{0.0438} \\ 
        4 & 0.3841 & 0.3683 & 4.2734  & 0.2609 \\ 
        5  & 0.3753 & 0.3686 & 1.8228 & 0.6385 \\ 
        6  & 0.3598 & 0.3223 & 11.6131 & \textbf{0.0060} \\ 
        7  & 0.4197 & 0.4121 & 1.8303 & 0.6115 \\ \hline
    \end{tabular}
\end{table}
% \begin{table}[h]
%     \centering
%     \caption{Pooled call listenership $L^{j}_{pooled}$ across all calls made in the respective time slot $j$ given by \ref{eq:timeslot_pooled}.}
%     \label{tab:timeslot_listenership}
%     \begin{tabular}{c|r|r|r|r}
%     	\hline
%     	Time Slot ID & Treatment & Control & \% listenership & p-value \\ \hline
%       	1  & 0.2043 & 0.1851 & 10.3437 & \textbf{0.0590} \\
%         2  & 0.1914 & 0.1788 & 7.0481 & 0.2052 \\ 
%         3  & 0.1915 & 0.1908 & 0.3722 & 0.9464 \\ 
%         4 & 0.1939 & 0.1974  & -1.7768 & 0.7412 \\ 
%         5  & 0.2072 & 0.1969 & 5.2434 & 0.3482 \\ 
%         6  & 0.1843 & 0.1687 & 9.2431 & 0.1269 \\ 
%         7  & 0.2114 & 0.2142 & -1.2890 & 0.8074 \\ \hline
%     \end{tabular}
% \end{table}
\\
We see a positive improvement in pick-up rate across all time slots and an improvement of 11.61\% in time slot id 6 especially. The p-values are significant for slots 3 and 6 at 0.05 level, and slot 1 at 0.10 level. We also include a similar analysis for the baseline phase in Table \ref{tab:timeslot_pickup_rates_preintervention} of the Appendix for the reader as well as additional analysis.

% \subsection{Overall pickup rates}
% The total improvement overall is $3.42\%$ improvement in pickup rate over random as detailed in Table \ref{tab:overall_pickup_rates}. Not only a statistical significant improvement is observed in the Treatment group post intervention, but we also see a statistical significant difference between the two groups post intervention highlighting the improvement in treatment group.

% \begin{table}[h]
%     \centering
%     \caption{Pooled call pickup rates across all calls made in the respective phases.}
%     \label{tab:overall_pickup_rates}
% \begin{tabular}{p{2cm}|p{2cm}|p{2cm}|p{2cm}}
% 	\hline
% 	Group & Pooled call pickup rate (Pre-) & Pooled call pickup rate (Post-) &  p-value \\ \hline
%   	Treatment  & 0.443 & \textbf{0.463} &\textbf{4.73e-07} \\
%     Control & 0.440 & 0.448 & 0.0428 \\ \hline
%   	p-value &  0.3369 &\textbf{0.0006} &  - \\ \hline

% \end{tabular}
% \end{table}

% \section{Tier-wise analysis for the baseline phase}
\begin{figure*}[]
  \centering
  \begin{tabular}{@{}c@{}}
    \includegraphics[width=.45\textwidth]{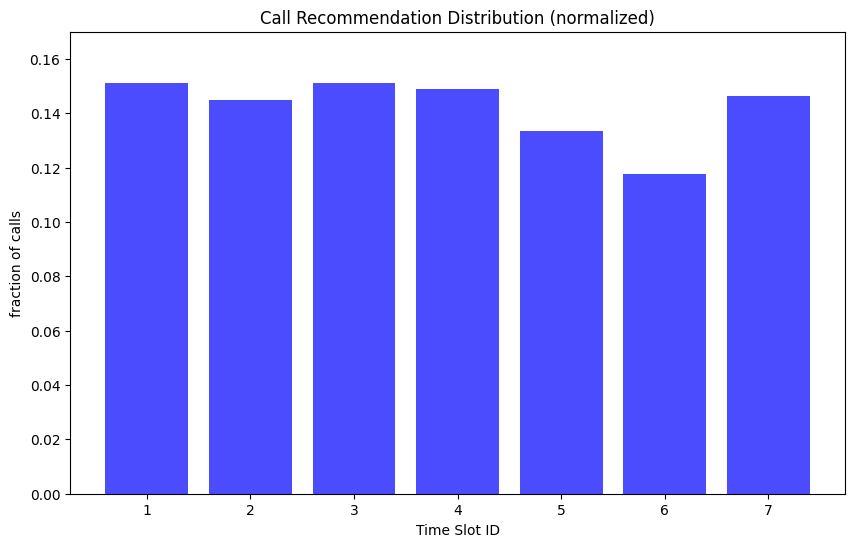} \\[\abovecaptionskip]
    \small \textbf{(a) Treatment Group}
  \end{tabular}
  \hspace{.7cm}
  \begin{tabular}{@{}c@{}}
    \includegraphics[width=.45\textwidth]{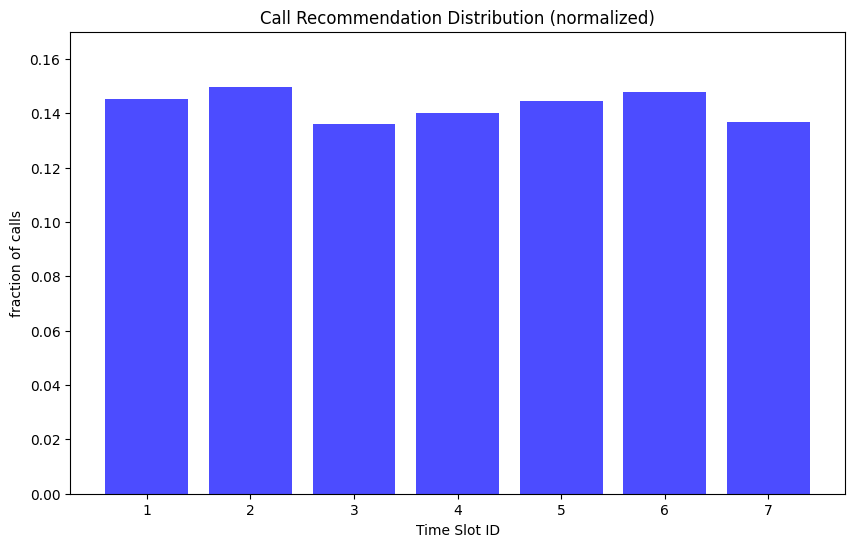} \\[\abovecaptionskip]
    \small \textbf{(b) Control Group}
  \end{tabular}
    \caption{The above bars represent the fraction of unique calls, i.e. the first call recommended by the algorithm without considering re-attempts. These give a truer representation of the call recommendation distribution or policy. (a) despite the distribution having a notable dip in slot 5 and 6, we still observe good pick-up success rates in slot 6 from Table \ref{tab:timeslot_pickup_rates}, and for (b) the distribution is almost uniform as expected.}
    \label{fig:call_dists}
\end{figure*}

\subsection{Call distributions}
Here we analyse the call distributions and how they changed during the different phases especially for the treatment group (collaborative bandits algorithm). The calls were also made according to a 3-2-2-2 pattern, i.e. if a user doesn't pick-up the first time we call them 2 more times with an interval of 5-10 minutes that day. If they still don't pick up, we call them twice the next day in the slot recommended by the algorithm for that day  -- the slot with the highest pick-up rate probability -- and twice the day after etc. until a call has been picked. If a call was picked in any one of these 9 attempts, the next call is made exactly a week after the first call. This pattern is identical to the call retry algorithm for the control group. We use the following method to obtain the call recommendation distribution $\pi(j)$ for time-slot $j$ for both the groups in the intervention phase,
\begin{align*}
    \texttt{count}(j) = \sum_{\forall i, t} A_{i, j, t, r=0}, &\qquad
    \pi(j) = \frac{\texttt{count}(j)}{\sum_{j=1}^{7} \texttt{count}(j)}.
\end{align*}
% If none of the 9 calls were picked, we assume the user to have dropped out and we don't call them again.

In Figure \ref{fig:call_dists}, we can see how the call recommendation distribution of the treatment group is not as uniform as that of the control group indicating that our algorithm is potentially finding the right time slots to call at while not heavily skewing towards any particular specific slots. This is important as it keeps the call load uniform.

\subsection{Off policy evaluation via importance sampling (IS)}

Deploying more algorithms in a field study would reduce the number of beneficiaries per baseline, which could potentially diminishing the significance of the findings. Due to these reasons, we decided to pick the baselines as detailed in the main paper.
One way to have an additional baseline without actually deploying an algorithm is by performing off-policy evaluation. We used a non-collaborative heuristic baseline that uses the pick-up matrix obtained from exploration phase of control group and then \textit{exploits} the slot with maximum pick-up rate per user during the intervention phase of the same group. We call this policy $q(i)$ where $i \in [1, 7]$ represents the time-slots. Let the set of calls be $C$, with $|C| = 25109$ (Table \ref{tab:call_volumes}). Then off-policy evaluation via IS, would be obtained by
$$V_q = \frac{1}{|C|} \sum_{c \in C} \frac{q(i_c)}{b(i_c)} r_c,$$
where $c$ is the call ID, $r_c$ is 1 if call $c$ was picked else 0, $i_c$ being corresponding slot, $q(i_c) = 1$ when $i_c$ is the slot with highest pick-up rate for that user else 0. Note that, $b(i_c) = 1/7 $ for the random behavioral policy. 
From this expression we obtain $V_q = 0.436$, which is less than the pick-up rate for the random policy $ 0.448$ (Table \ref{tab:overall_pickup_rates_active}). We further confirm the statistical significance of this results by boot-strapping over a subset of beneficiaries. This indicates that a longer exploration period is needed for per user inference, establishing the strength of a collaborative policy that had a pick-up rate of $0.463$ (Table \ref{tab:overall_pickup_rates_active}), which reduces the need for exploration significantly.

\subsection{Summary}
Our paper provided results of a field study and followed with a tiered analysis for comparison of collaborative bandits against a random calling strategy in the field. This is the first such field study and tiered analysis of the Kilkari program in India, the largest maternal mobile health program in the world. 

The tiered analysis reveals that the collaborative bandit algorithm (the treatment group) significantly improved call pick-up rates compared to the random calling strategy (random control group), particularly for beneficiaries in the middle and bottom tiers. This demonstrates the algorithm's effectiveness in optimizing call scheduling and enhancing message delivery within the Kilkari program.

Some of the generalizable insights learned from this work are as follows. 
Collaborative bandit algorithms offer a promising approach for personalized intervention delivery in mobile health By iteratively learning from user responses and interactions, these algorithms can adapt to individual preferences and maximize engagement as compared to random calling algorithms.
% In this study, we implemented a collaborative bandit algorithm within the Kilkari platform and conducted a field trial involving approximately $6500$ beneficiaries. 
Our goal was to evaluate the algorithm's ability to improve call pick-up rates compared to the baseline random calling strategy.
This research contributes to the growing body of literature in the application of machine learning in mobile health interventions \cite{verma2023deployed, mate2022field, nair2022adviser}. 
By demonstrating the effectiveness of a collaborative bandit algorithm in a real-world setting, we highlight the potential for personalized call scheduling to enhance the reach and impact of maternal health programs at scale.
Given the national scope of Kilkari and the potential for improved message delivery to millions of beneficiaries, our findings have significant implications for public health policy and practice in India and beyond.

%%%%%%%%%%%%%%%%%%%%%%%%%%%%%%%%%%%%%%%%%%%%%%%%%%%%%%%%%%%%%%%%%%%%%%%%

%%% The acknowledgments section is defined using the "acks" environment
%%% (rather than an unnumbered section). The use of this environment 
%%% ensures the proper identification of the section in the article 
%%% metadata as well as the consistent spelling of the heading.

% \begin{acks}
% If you wish to include any acknowledgments in your paper (e.g., to 
% people or funding agencies), please do so using the `\texttt{acks}' 
% environment. Note that the text of your acknowledgments will be omitted
% if you compile your document with the `\texttt{anonymous}' option.
% \end{acks}

%%%%%%%%%%%%%%%%%%%%%%%%%%%%%%%%%%%%%%%%%%%%%%%%%%%%%%%%%%%%%%%%%%%%%%%%

%%% The next two lines define, first, the bibliography style to be 
%%% applied, and, second, the bibliography file to be used.

\bibliography{sample}

%%%%%%%%%%%%%%%%%%%%%%%%%%%%%%%%%%%%%%%%%%%%%%%%%%%%%%%%%%%%%%%%%%%%%%%%
\newpage

\appendix
\section*{Appendix}
We present a similar tier-wise analysis as demonstrated in Table \ref{tab:timeslot_pickup_rates} in Table \ref{tab:timeslot_pickup_rates_preintervention} but for the baseline phase. The results tell us that the \% pick-up rate difference is greater and more significant during the intervention phase. We also provide a Difference in Difference (DiD) number in the final column, which is calculated as follows,
\begin{align}
\left( PR_{\text{pooled}}^{\text{Treatment, Intervention}} - PR_{\text{pooled}}^{\text{Treatment, Baseline}} \right) - 
\left( PR_{\text{pooled}}^{\text{Control, Intervention}} - PR_{\text{pooled}}^{\text{Control, Baseline}} \right).
\end{align}
The \textit{positive} DiD values in Table \ref{tab:timeslot_pickup_rates_preintervention} also indicate that the improvements observed across the two groups are larger in the intervention phase than the baseline phase, further strengthening our claim.

\begin{table}[h]
    \centering
    \caption{Pooled call pick-up rates $PR^{active, j}_{pooled}$ across all calls made in the respective time slot $j$ given by \ref{eq:timeslot_pooled} during the baseline phase for only the active users.}
    \label{tab:timeslot_pickup_rates_preintervention}
    \begin{tabular}{c|r|r|r|r|r}
    	\hline
    	Time Slot ID & Treatment & Control & \% pick-up rate & p-value & DiD\\ \hline
       	1  & 0.3690 & 0.3610 & 2.2024 & 0.4387 & 0.0168 \\
        2  & 0.3646 & 0.3810 & -4.2999 & 0.1153 & 0.0308 \\ 
        3  & 0.4034 & 0.4072 & -0.9454 & 0.7185 & 0.0322 \\ 
        4 & 0.4131 & 0.3979 & 3.8158  & 0.1704 & 0.0006 \\ 
        5  & 0.4016 & 0.3961 & 1.3702 & 0.6309 & 0.0013 \\ 
        6  & 0.3846 & 0.3741 & 2.8050 & 0.3606 & 0.0269 \\ 
        7  & 0.4678 & 0.4610 & 1.4862 & 0.5777 & 0.0007 \\ \hline
    \end{tabular}
\end{table}
\begin{table}[h]
    \centering
    \caption{Time slot IDs mapping to the corresponding time windows in 24 hr format.}
    \label{tab:timeslot_ids_details}
    \begin{tabular}{c|r|r}
        \hline
        Time Slot ID & Start time & End time \\ \hline
        1  & 06:45:00 & 08:45:00 \\
        3  & 08:45:00 & 10:45:00  \\ 
        2  & 10:45:00 & 12:45:00  \\ 
        4 & 12:45:00 & 14:45:00  \\ 
        5  & 14:45:00 & 16:45:00  \\ 
        6  & 16:45:00 & 18:45:00  \\ 
        7  & 18:45:00 & 20:45:00  \\ \hline
    \end{tabular}
\end{table}

\end{document}